\documentclass[letterpaper]{article} % DO NOT CHANGE THIS
\usepackage{aaai23}  % DO NOT CHANGE THIS
\usepackage{times}  % DO NOT CHANGE THIS
\usepackage{helvet}  % DO NOT CHANGE THIS
\usepackage{courier}  % DO NOT CHANGE THIS
\usepackage[hyphens]{url}  % DO NOT CHANGE THIS
\usepackage{graphicx} % DO NOT CHANGE THIS
\usepackage{natbib}  % DO NOT CHANGE THIS AND DO NOT ADD ANY OPTIONS TO IT
\usepackage{caption} % DO NOT CHANGE THIS AND DO NOT ADD ANY OPTIONS TO IT
\usepackage{algorithm}
\usepackage{algorithmic}

\usepackage[table,xcdraw]{xcolor} 
\usepackage{color}
\usepackage{multirow}
\usepackage{arydshln}
\usepackage{graphicx}
\usepackage{float}
\usepackage{subfig}
\usepackage{booktabs}
\usepackage{tabularx}

\usepackage{amsmath}
\usepackage{amssymb}
\usepackage{xspace}
\usepackage{tablefootnote}
\usepackage{colortbl}
\definecolor{mygray}{gray}{.9}
\definecolor{LightYellow}{rgb}{0.99,0.94,0.7}
\usepackage{newfloat}
\usepackage{listings}
\DeclareCaptionStyle{ruled}{labelfont=normalfont,labelsep=colon,strut=off} % DO NOT CHANGE THIS
\floatstyle{ruled}
\newfloat{listing}{tb}{lst}{}
\floatname{listing}{Listing}
\title{DeMT: Deformable Mixer Transformer for Multi-Task Learning of Dense Prediction}
\author {
    Yangyang Xu \textsuperscript{\rm 1},
    Yibo Yang  \textsuperscript{\rm 3},
    Lefei Zhang \textsuperscript{\rm 1,2}\thanks{Corresponding Author.}
}
\affiliations {
    \textsuperscript{\rm 1} National Engineering Research Center for Multimedia Software, School of Computer Science, Wuhan University, China\\
    \textsuperscript{\rm 2} Hubei Luojia Laboratory, China\quad
    \textsuperscript{\rm 3} JD Explore Academy, China\\
     \{yangyangxu, zhanglefei\}@whu.edu.cn, \  ibo@pku.edu.cn
}

\usepackage{bibentry}
\begin{document}

\maketitle

\begin{abstract}
Convolution neural networks (CNNs) and Transformers have their own advantages and both have been widely used for dense prediction in multi-task learning (MTL). Most of the current studies on MTL solely rely on CNN or Transformer. 
In this work, we present a novel MTL model by combining both merits of deformable CNN and query-based Transformer for multi-task learning of dense prediction.
Our method, named DeMT, is based on a simple and effective encoder-decoder architecture ($i.e.,$ deformable mixer encoder and task-aware transformer decoder).
First, the deformable mixer encoder contains two types of operators: the channel-aware mixing operator leveraged to allow communication among different channels ($i.e.,$ efficient channel location mixing), and the spatial-aware deformable operator with deformable convolution applied to efficiently sample more informative spatial locations ($i.e.,$ deformed features). 
Second, the task-aware transformer decoder consists of the task interaction block and task query block. 
The former is applied to capture task interaction features via self-attention.
The latter leverages the deformed features and task-interacted features to generate the corresponding task-specific feature through a query-based Transformer for corresponding task predictions.
Extensive experiments on two dense image prediction datasets, NYUD-v2 and PASCAL-Context, demonstrate that our model uses fewer GFLOPs and significantly outperforms current Transformer- and CNN-based competitive models on a variety of metrics.
The code are available at https://github.com/yangyangxu0/DeMT.
\end{abstract}

\section{Introduction}
\label{sec:intro}
Human vision capability is powerful and can perform different tasks from one visual scene, such as classification, segmentation, recognition, etc.
Therefore, multi-task learning (MTL) research is topical in computer vision.
We expect to develop a powerful vision model to do multiple tasks simultaneously in different visual scenarios, and this model is expected to work efficiently.
As shown in Figure 1, in this paper, we aim to develop a powerful vision model to learn multiple tasks, including semantic segmentation, human parts segmentation, depth estimation, boundary detection, saliency estimation, and normal estimation simultaneously, and this model is expected to work efficiently.

Recently, existing works~\cite{MTL_attn_2019,Mti-net_2020,mtlP_2021,multitask_mtst_2021,atrc_2021,mqtrans_xy,transfmult2022} have adopted CNN and Transformer technologies to advance the MTL of dense prediction.
Although CNN-based MTL models are carefully proposed to achieve promising performance on the multi-task dense prediction task, these models still suffer from the limitations of convolutional operations, $i.e.,$ lacking global modeling and cross-task interaction capability.
Some works~\cite{atrc_2021,Mti-net_2020} develop a distillation scheme to increase the expressiveness of the cross-task and global information passing via enlarging the receptive field and stacking multiple convolutional layers but still cannot build global dependency directly. 
For modeling global and cross-task interaction information, Transformer-based MTL models~\cite{transfmult2022,mqtrans_xy} utilize the efficient attention mechanism~\cite{attent2017} for global modeling and task interactions. 
However, such a self-attention approach may fail to focus on task awareness features because the \textit{queries}, \textit{keys} and \textit{values} are based on the same feature.
Regular self-attention may lead to high computational costs and limit the ability to disentangle task-specific features.

We can see that the CNN-based models better capture the multiple task context in a local field but suffer from a lack of global modeling and task interaction. The Transformer-based models better focus on global information of different tasks. However, they ignore task awareness and introduce many computation costs. Therefore, a technical challenge in developing a better MTL model is how to combine the merits of CNN-based and Transformer-based MTL models.

To address the challenges, we introduce the \textbf{De}formable \textbf{M}ixer \textbf{T}ransformers (DeMT): a simple and effective method for multi-task dense prediction based on combining both merits of deformable CNN and query-based Transformer.
Specifically, our DeMT consists of the deformable mixer encoder and task-aware transformer decoder.
Motivated by the success of deformable convolutional networks~\cite{defcnn_2019} in vision tasks,
our deformable mixer encoder learns different deformed features for each task based on more efficient sampling spatial locations and channel location mixing ($i.e.,$ deformed feature).
It learns multiple deformed features highlighting more informative regions with respect to the different tasks. 
In the task-aware transformer decoder, the multiple deformed features are fused and fed into our task interaction block. 
We use the fused feature to generate task-interacted features via a multi-head self-attention for model task interactions.
To focus on the task awareness of each individual task, we use deformed features directly as \textit{query} tokens.
We expect the set of candidate \textit{key/value} to be from task-interacted features.
Then, our task query block tasks the deformed features and task-interacted features as input and generates the task awareness features.
In this way, our deformable mixer encoder selects more valuable regions as deformed features to alleviate the lack of global modeling in CNN.
The task-aware transformer decoder performs the task interactions by self-attention and enhances task awareness via a query-based Transformer. This design both reduces computational costs and focuses on task awareness features.  
Through extensive experiments on several publicly MTL dense prediction datasets, we demonstrate that the proposed DeMT method achieves state-of-the-art results on a variety of metrics.

The contributions of this paper are as follows:
1) We propose a simple and effective DeMT method for MTL of dense prediction via combining both merits of CNN and Transformer.
Most importantly, our approach not only alleviates the lack of global modeling in MTL models using CNNs but also avoids the lack of task awareness in MTL models using Transformers.
2) We introduce a deformable mixer transformer (DeMT) model which consists of the deformable mixer encoder (Section~\ref{subsec:enc}) and task-aware transformer decoder (Section~\ref{subsec:dec}).
The deformable mixer encoder produces the deformed features.
The task-aware transformer decoder uses the deformed features to model task interaction via a self-attention and focus on the task awareness features via a query-based transformer.
3) The extensive experiments on NYUD-v2~\cite{NYUD2012} and PASCAL-Context~\cite{pascal2014} and visualization results show the efficacy of our model.
DeMT's strong performance on MTL can demonstrate the benefits of combining the deformable CNN and query-based Transformer. 

\section{Related Work}
\label{sec:relate}

\begin{figure*}[!t]
\centering
  \includegraphics[width=0.95\textwidth]{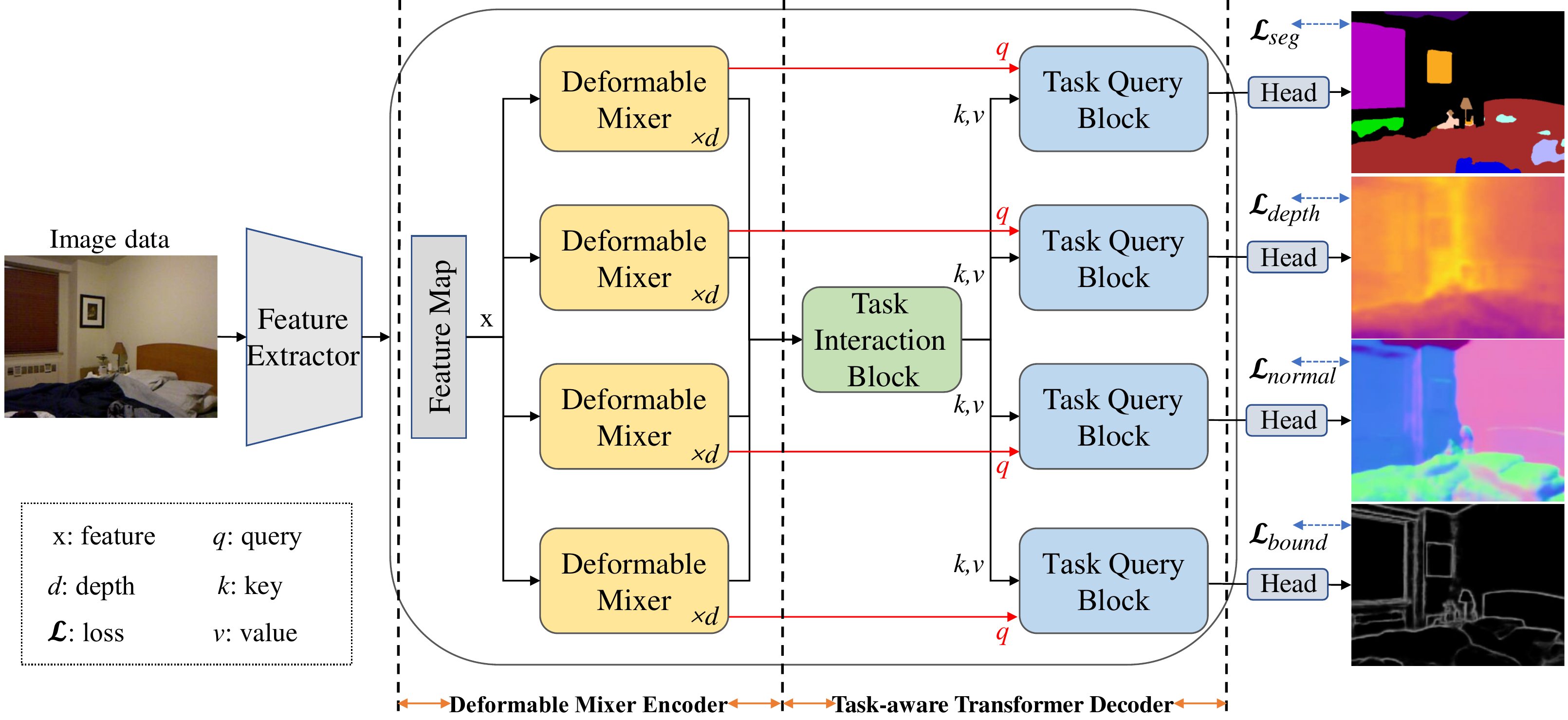}
  \caption{An overview of our model jointly handles multiple tasks with a unified encoder-decoder architecture. Our DeMT model consists of the deformable mixer encoder and task-aware transformer decoder. The depth $d$ is the number of repetitions of the Deformable Mixer (ablation on the $d$ in Table~\ref{tab:depth}).}
  \label{fig:over_view}
\end{figure*}

\subsection{Multi-Task Learning (MTL)}
MTL has dramatically evolved with the development of Deep Neural Networks and Vision Transformers.
MTL tasks are mainly distributed in two aspects: model structures~\cite{atrc_2021} and task loss weighting optimization~\cite{mtl_loss2021}.
In the vision domain, the core idea of MTL is to use a single model to predict semantic segmentation, human parts segmentation, depth, surface normal, boundary, $etc.,$ which is an interesting topic. 
MuST~\cite{multitask_mtst_2021} model uses the knowledge in independent specialized teacher models to train a general model for creating general visual representations.
Several recent MTL frameworks follow different technologies:~\cite{NDDR-CNN_2019,Mti-net_2020} is CNN-based MTL model,~\cite{atrc_2021} is Neural Architecture Search (NAS)-based model and~\cite{transfmult2022,mqtrans_xy,xu2022mtformer} are Transformer-based models.
\cite{MTL_survey_2021} states that the MTL structures in vision tasks can be summarized into two categories: encoder- and decoder-focused architectures.
Some encoder-focused works~\cite{encoder_base_2018,encoder_base2_2018} rely on a shared encoder to learn a general visual feature. The features from the shared encoder are input into the task-specific heads to perform the dense predictions for every task.
The decoder-focused works~\cite{NDDR-CNN_2019,PAP-Net_2019,atrc_2021} use a shared backbone network to extract a shared feature for each task. Then, the designed task-specific module captures valuable information from the shared feature.
However, these MTL methods primarily focus on the shared feature, which is hard to disentangle the task-specific features.

\subsection{Deformable CNNs and Transformers.}
\textbf{Deformable CNNs.}
Deformable ConvNets~\cite{deformablecnn2017,defcnn_2019} harness the enriched modeling capability of CNNs~\cite{yang2020sognet} via deformable convolution and deformable spatial locations. Deformable ConvNets is the first to achieve competitive results in vision tasks ($e.g.,$ object detection and semantic segmentation, $etc.$) using deformable convolution.
Deformable transformers~\cite{deformableDetr,deformatten2022} attend to a small set of crucial sampling points around a reference and capture more informative features.

\noindent
\textbf{Transformers.}
Transformer and attention~\cite{attent2017} mechanism models were first employed in natural language processing (NLP) tasks with good performance.
Recently, the transformer structures have also produced impressive results in computer vision and tend to replace CNN progressively. ViT~\cite{ViT2021} is the first work to derive from the attention mechanism for computer vision tasks. More Transformer-based approaches~\cite{detr_2020,swin,Ranftl2021,densePredic_pvt_2021,lan2022siamese,ru2022} have been introduced by improving the attention mechanism for dense prediction tasks.
Recently, these works are also extended to the MTL domain to learn good representations for multiple task predictions.
In contrast, we find the deformed features to focus on the valuable region for different tasks.
In addition, we use the query-based transformer approach for modeling and leverage deformed features as queries in transformer calculations to enhance task-relevant features.
These queries can naturally disentangle the task-specific feature from the fused feature.
Our approach combines the respective advantages of CNN and Transformer, achieving state-of-the-art on NYUD-v2 and PASCAL-Context datasets.

\section{The DeMT Method}
\label{sec:method}

\subsection{Overall Architecture}
We describe the overall framework of our architecture in Figure~\ref{fig:over_view}.
DeMT is the result of a non-shared encoder-decoder procedure: First, we design a deformable mixer encoder to encode task-specific spatial features for each task.
Second, the task interaction block and task query block are proposed to model the decode the task interaction information and decode task-specific features via self-attention.
In the following section, we describe our task losses.

\subsection{Feature Extractor}
\label{sec:fe}
The feature extractor is utilized to aggregate multi-scale features and manufacture a shared feature map for each task. 
The initial image data $X_{in} \in \mathbb{R}^{H \times W \times 3}$ (3 means image channel) is input to the backbone, which then generates four stages of image features.
Then the four stage image features are up-sampled to the same resolution, and then they are concatenated along the channel dimension to obtain an image feature $X \in \mathbb{R}^{\frac{H}{4} \times \frac{W}{4} \times C}$, where $H$, $W$, and $C$ are the height, width, and channel of the image feature, respectively. 

\subsection{Deformable Mixer Encoder}
\label{subsec:enc}
\begin{figure}[!t]
\centering
  \includegraphics[width=0.47\textwidth]{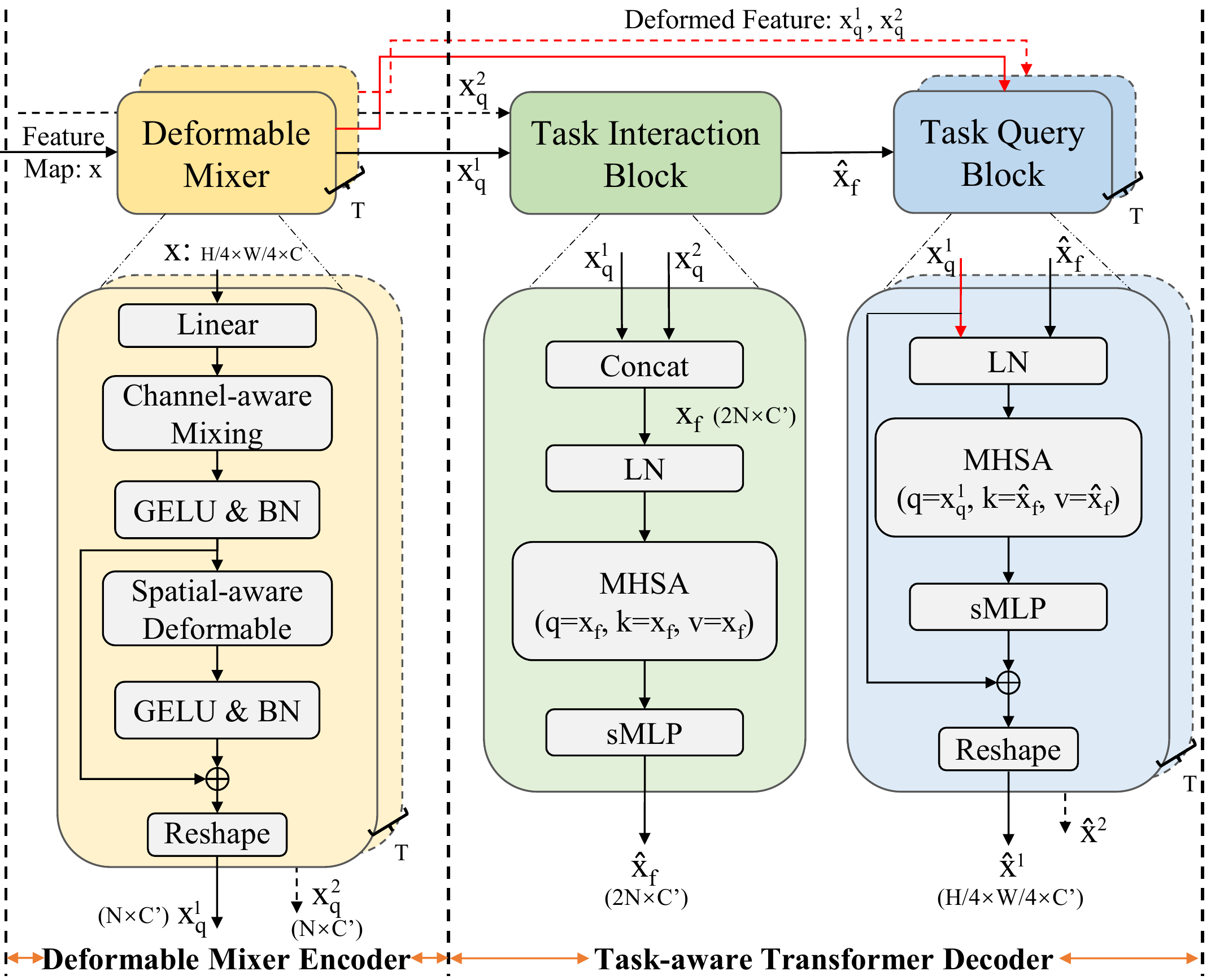}
  \caption{Illustration of our DeMT components. For simplicity, we assume there are two tasks (T=2) in this figure. Small MLP (sMLP) only consists of Linear and LayerNorm functions.}
  \label{fig:models}
\end{figure}

\textbf{The motivation.} Inspired by the success of the Deformable ConvNets~\cite{defcnn_2019} and Deformable DETR models~\cite{deformableDetr}, we propose the deformable mixer encoder that adaptively provides more efficient receptive fields and sampling spatial locations for each task.
For this purpose, the deformable mixer encoder is designed to separate the mixing of spatial-aware deformable spatial features and channel-aware location features.
As shown in Figure~\ref{fig:models} (left), the spatial-aware deformable and channel-aware mixing operators are interleaved to enable interaction of both input feature dimensions ($HW \times C$).

Specifically, we propose a deformable mixer encoder to capture the unique receptive regions corresponding to the individual task.
The deformable mixer only attends to a small set of crucial sampling points which are learnable offset.
The spatial-aware deformable is capable of modeling spatial context aggregation. 
Then the spatial-aware deformable, channel-aware mixing, and layer normalization operators are stacked to form one deformable mixer. The effect of the depth of the deformable mixer stack on the model is shown in the Table~\ref{tab:depth} ablation experiment.

The deformable mixer encoder structure is shown in Figure~\ref{fig:models}.
First, a linear layer reduces the channel dimension of the image feature $X\in \mathbb{R}^{\frac{H}{4} \times \frac{W}{4} \times C}$ from $C$ to a smaller dimension $C'$.  
The linear layer can be written as follows:
\begin{equation}
  \begin{aligned}
    \begin{array}{ll}
        {X} = {W}\cdot \operatorname{Norm}({X}),\label{eq:linear}
    \end{array}
  \end{aligned}
\end{equation}
where Norm means LayerNorm function. After the linear layer, we obtain a smaller dimension image feature map $X\in \mathbb{R}^{\frac{H}{4} \times \frac{W}{4} \times C'}$ as the input for the downstream.

\noindent
\textbf{Channel-aware mixing.}
The channel-aware mixing allows communication between different channels.
The channel-aware mixing applies the standard point-wise convolution (the convolving kernel is 1$\times$1) to mix channel locations.
It can be formulated as:
\begin{equation}
X_{C'} = \sum_{C'=0}^{C'-1} W_1 \cdot{X_{C'}} + b,
\end{equation}
where the $W_1$ is the point-wise convolution weight. $b$ is a learnable bias. Subsequently, we add GELU activation and BatchNorm as well. This operation is calculated as:
\begin{equation}\label{eq:bn1}
 X_{C'} =  \operatorname{BN}(\sigma(X_{C'})),
\end{equation}
where $\sigma(\cdot)$ is the non-linearity function (GELU); BN is the BatchNorm operation.

\noindent
\textbf{Spatial-aware deformable.}
Given the input image feature $X_{i,j}\in \mathbb{R}^{\frac{H}{4} \times \frac{W}{4} \times C'}$ from Eq.(\ref{eq:bn1}), the point $(i,j)$ is the spatial location on the single channel.

To generate the relative offsets with respect to the reference point, the image feature $X_{C'}$ is fed to the convolution operator to learn the corresponding offsets $\Delta_{(i,j)}$ for all reference points.
For each location point $(i,j)$ on the image feature $X$, the spatial deformable can be written as:
\begin{equation}
D_S(X_{i,j}) = \sum_{C'=0}^{C'-1}  W_2 \cdot {X}((i,j)+\Delta_{(i,j)}, C'),
\end{equation}
where the $W_2$ is a deformable weight. The $\Delta_{(i,j)}$ is the learnable offset. 
The spatial-aware deformable is followed by a GELU activation, BatchNorm, and residual connection:
\begin{equation}\label{eq:encoder}
 X_q = \text{Reshape}(X_{C'} + \operatorname{BN}(\sigma(D_S(X_{i,j})))),
\end{equation}
where the \textit{Reshape} is applied to flatten the feature $X_q\in \mathbb{R}^{\frac{H}{4} \times \frac{W}{4} \times C'}$ to a sequence $\mathbb{R}^{N\times C'}$ ($N=\frac{H}{4} \times \frac{W}{4} $).
When there are $T$ tasks, the deformable mixer encoder generate a feature set {($X^1_q, X^2_q, \cdots X^T_q)$} ($T$ means task number)  (See Figure~\ref{fig:models}).
These output task-specific features are learned by a deformable mixer that we refer to as \textit{deformed features}, which we add to the input of the downstream blocks (task interaction block and task query block).

\subsection{Task-aware Transformer Decoder}
\label{subsec:dec}
In the task-aware transformer decoder, we design the task interaction block and task query block (See Figure~\ref{fig:models}). 
It is important for MTL to consider task interactions. Thus, we propose a task interaction block to capture the task interactions at every task via an attention mechanism. Each task interaction block is composed of two parts, i.e., a multi-head self-attention module (MHSA) and a small Multi-Layer Perceptron (sMLP). The downstream task query block also consists of the MHSA and the sMLP. 
The difference between the task interaction block and the task query block is that their query features are fundamentally different.
The feature is projected into the queries (\textit{Q}), keys (\textit{K}) and values (\textit{V}) of dimension $d_k$ and self attention is being computed by the \textit{Q}, \textit{K} and \textit{V}.
The self-attention operator is calculated as:
\begin{align}\label{eq:mhsa}
    &\text{MHSA}(Q, K, V) = \text{softmax}(\frac{QK^{T}}{\sqrt{d_k}})V,
\end{align}
where $Q \in \mathbb{R}^{N \times C'}$, $K \in  \mathbb{R}^{N \times C'}$ and $ V \in  \mathbb{R}^{N \times C'}$ are the query, key and value matrices; $\text{MHSA}(Q, K, V) \in \mathbb{R}^{N \times C'}$.

\noindent
\textbf{Task interaction block.}
As illustrated in Figure~\ref{fig:models} (center), We first concatenate the deformed features from the deformable mixer encoder output.

\begin{equation}
 X_f =  \operatorname{Concat}(X^1_q, X^2_q, \cdots X^T_q),
\end{equation}
where $X_f \in \mathbb{R}^{TN \times C'}$ is the fused feature. The $T$ means task number in $X^T_q \in \mathbb{R}^{N \times C'}$.
Then, for efficient task interaction, we construct a self-attention strategy via the fused feature $X_f$:
\begin{align}\label{eq:mhsafusion}
    X_f' =\ &\text{MHSA}(Q=\operatorname{LN}(X_f), K=\operatorname{LN}(X_f), V=\operatorname{LN}(X_f)),\\
    \hat X_f =\ &\operatorname{sMLP}(X_f'),
\end{align}
where $\hat X_f \in \mathbb{R}^{TN \times C'}$ is the task-interacted feature. LN means LayerNorm function. sMLP consists of a linear layer and a LayerNorm.

\noindent
\textbf{Task query block.}
As illustrated in Figure~\ref{fig:models} (right), we take the deformed feature $X_q$ as task query and the task-interacted feature $\hat X_f$ as key \& value to MHSA.
The deformed feature is applied as a query in MHSA to decode the task awareness feature from the task-interacted feature for each task prediction.
We first apply the LayerNorm in parallel to generate queries $Q$, keys $K$ and values $V$:

\begin{align}\label{eq:taskqueryln}
   \hat Q=\operatorname{LN}(X_q),\quad
   \hat K=\operatorname{LN}(\hat X_f),\quad
   \hat V=\operatorname{LN}(\hat X_f),
\end{align}
where LN is the layer normalization. $X_q$ and $\hat X_f$ are the output of deformable mixer encoder and task interaction block, respectively. 
Then, the task query block operation using a MHSA is calculated as:
\begin{align}\label{eq:taskquery}
    \hat X_q =\ &\text{MHSA}(\hat Q, \hat K, \hat V),\\
    \hat{X} =\ &\text{Reshape}(X_q + \operatorname{sMLP}(\hat X_q)),
\end{align}
where the residual feature $X_q$ comes from Eq.~(\ref{eq:encoder}).
The task awareness feature $\hat{X} \in \mathbb{R}^{\frac{H}{4} \times \frac{W}{4} \times C'}$ is reshaped from $\mathbb{R}^{N \times C'}$ ($N=\frac{H}{4} \times \frac{W}{4} $) via \textit{Reshape} operation.

\subsection{Loss Function}
For balancing the loss contribution for each task, we set the weight $\alpha_t$ to decide the loss contribution for the task $t$.
A weighted sum $\mathcal L_{total}$ of task-specific losses:
	\begin{equation}\label{equ:loss}
	\begin{split}
	    \begin{aligned}
    {\mathcal L_{total}} &= \sum_{t=1}^{T} \alpha_{t}{\mathcal L_{t}}, \\
       \end{aligned}
    \end{split}
	\end{equation}
where the $\mathcal L_{t}$ is a loss function for task $t$. For fair comparisons, we use $\alpha_t$ and $\mathcal L_{t}$  consistent with ATRC~\cite{atrc_2021} and MQTransformer~\cite{mqtrans_xy}.

% table 1 NYUD
\begin{table*}[!tp]
\small
\centering
\caption{Comparison of the MTL models with state-of-the-art on NYUD-v2 dataset. The notation '$\downarrow$': lower is better. The notation '$\uparrow$': higher is better. $\Delta_m$ denotes average per-task performance drop. "Params" denotes parameters.}
\label{tab:stoa_nyud_v2}
		\begin{tabular}{llllllllr}
\hline\noalign{\smallskip}
 \multirow{2}*{Model}  & \multirow{2}*{Backbone}  &Params  &GFLOPs  & {SemSeg}   &Depth & Normal & Bound &\multirow{2}*{$\Delta_m[\%]$$\uparrow$}\\
      &  &(M) &(G) &(IoU)$\uparrow$   &(rmse)$\downarrow$   & (mErr)$\downarrow$  &(odsF)$\uparrow$\\
%\noalign{\smallskip}
\hline
%\noalign{\smallskip}
single task baseline &HRNet18   &16.09 &40.93  &38.02 &0.6104 &20.94 &76.22 &0.00\\
multi-task baseline  &HRNet18   &4.52  &17.59  &36.35 &0.6284 &21.02 &76.36 &-1.89\\
%\noalign{\smallskip}
%\hdashline
%\noalign{\smallskip}
Cross-Stitch\cite{Cross-Stitch_2016}  &HRNet18   &4.52  &17.59  &36.34  &0.6290  &20.88  &76.38 &-1.75\\
Pad-Net\cite{Pad-net_2018}            &HRNet18   &5.02  &25.18  &36.70  &0.6264  &20.85  &76.50 &-1.33\\
PAP\cite{PAP-Net_2019}                &HRNet18   &4.54  &53.04  &36.72  &0.6178  &20.82  &76.42 &-0.95\\
PSD\cite{pattern_struct_diffusion_2020} &HRNet18 &4.71  &21.10  &36.69  &0.6246  &20.87  &76.42 &-1.30\\
NDDR-CNN\cite{NDDR-CNN_2019}          &HRNet18   &4.59  &18.68  &36.72  &0.6288  &20.89  &76.32 &-1.51\\
MTI-Net\cite{Mti-net_2020}            &HRNet18   &12.56 &19.14  &36.61  &0.6270  &20.85  &76.38 &-1.44\\
ATRC\cite{atrc_2021}                  &HRNet18   &5.06  &25.76  &38.90  &0.6010  &20.48  &76.34 &1.56\\
\rowcolor{mygray}DeMT (Ours)          &HRNet18  &4.76   &22.07  &39.18  &0.5922  &20.21  &76.40 &2.37\\

%\noalign{\smallskip}
\hdashline
%\noalign{\smallskip}
single task baseline            &Swin-T  &115.08 &161.25  &42.92   &0.6104  &20.94  &76.22 &0.00\\
multi-task baseline             &Swin-T  &32.50  &96.29   &38.78   &0.6312  &21.05  &75.60 &-3.74\\
MQTransformer\cite{mqtrans_xy}  &Swin-T  &35.35  &106.02  &43.61   &0.5979  &20.05  &76.20 &0.31\\
\rowcolor{mygray}DeMT (Ours)    &Swin-T  &32.07  &100.70  &46.36   &0.5871  &20.65  &76.90 &3.36\\

%\noalign{\smallskip}
\hdashline      
%\noalign{\smallskip}
single task baseline           &Swin-S  &200.33 &242.63 &48.92  &0.5804  &20.94  &77.20  &0.00\\
multi-task baseline            &Swin-S  &53.82  &116.63 &47.90  &0.6053  &21.17  &76.90  &-1.96\\
MQTransformer\cite{mqtrans_xy} &Swin-S  &56.67  &126.37 &49.18  &0.5785  &20.81  &77.00  &1.59\\
\rowcolor{mygray}DeMT (Ours)   &Swin-S  &53.03  &121.05 &51.50  &0.5474  &20.02  &78.10  &4.12\\
\hline
  \end{tabular}
\end{table*}

\section{Experiment}
\label{sec:exp}
In this section, we conduct extensive experiments on two widely-used dense prediction datasets to evaluate the performance of our method on different metrics. We also show the visualization results on different datasets.

\subsection{Experimental Setup}
\noindent
\textbf{Implementation.}
All the leveraged backbones generate four scales (1/4, 1/8, 1/16, 1/32) features to perform multi-scale aggregation in our feature extractor (Section~\ref{sec:fe}).
We train our model with SGD setting the learning rate to $10^{-3}$ and weight decay to $5\times10^{-4}$.
The whole experiments are performed with pre-trained models on ImageNet.
All our experiments are performed on the Pytorch platform with eight A100 SXM4 40GB GPUs.

\noindent
\textbf{Datasets.}
We conduct experiments on two publicly accessible datasets, NYUD-v2~\cite{NYUD2012} and PASCAL-Context~\cite{pascal2014}. 
NYUD-V2 is comprised of pairs of RGB and Depth frames that 795 images are used for training and 654 images for testing.
NYUD-V2 usually is mainly adopted for semantic segmentation (‘SemSeg’), depth estimation (‘Depth’), surface normal estimation (‘Normal’), and boundary detection (‘Bound’) tasks by providing dense labels for every image.
PASCAL-Context training and validation contain 10103 images, while testing contains 9637 images. 
PASCAL-Context usually is adopted for semantic segmentation ('SemSeg'), human parts segmentation ('PartSeg'), saliency estimation ('Sal'), surface normal estimation ('Normal'), and boundary detection ('Bound') tasks by providing annotations for the whole scene.

\noindent
\textbf{Metrics.}
We adopt five evaluation metrics to compare our model with other prior multi-task models: mean Intersection over Union (mIoU), root mean square error (rmse), mean Error (mErr), optimal dataset scale F-measure (odsF), and maximum F-measure (maxF).
The average per-task performance drop ($\Delta_m$) is used to quantify multi-task performance. $\Delta_m=\frac{1}{T} \sum_{i=1}^T(F_{m,i}-F_{s,i})/F_{s,i}\times100\%$, where $m$, $s$ and $T$ mean multi-task model, single task baseline and task numbers.
$\Delta_m$: the higher is the better.

\noindent
\textbf{Backbones.}
We test our method using several CNN and Vision Transformer backbones: HRNetV2-W18-small (HRNet18), HRNetV2-W48 (HRNet48)~\cite{HRnet_19}, Swin-Tiny (Swin-T), Swin-Small (Swin-S) and Swin-Base (Swin-B)~\cite{swin}.

\subsection{Comparison with the state-of-the-art}
We compare our model with CNN-based and Transformer-based models to show the advantages of our method.

% table2 PASCAL-Context
\begin{table*}[!ht]
\small
\centering
\caption{Comparison of the MTL models with state-of-the-art on PASCAL-Context dataset. The notation ‘$\downarrow$’: lower is better. The notation ‘$\uparrow$’: higher is better. $\Delta_m$ denotes average per-task performance drop (the higher is the better).}
%\vspace{-1.5mm}
\label{tab:stoa_pascal}
%\resizebox{1.08\columnwidth}{!}{
\begin{tabular}{lllllllr}
\hline\noalign{\smallskip}
 \multirow{2}*{Model}  & \multirow{2}*{Backbone}  & {SemSeg}   &PartSeg &Sal  & Normal & Bound &\multirow{2}*{$\Delta_m[\%]$$\uparrow$}\\
     &  & (IoU)$\uparrow$  & (IoU)$\uparrow$  &(maxF)$\uparrow$  &(mErr)$\downarrow$  &(odsF)$\uparrow$\\
%\noalign{\smallskip}
\hline
%\noalign{\smallskip}
single task baseline  &HRNet18   &62.23  &61.66 &85.08  &13.69 &73.06 &0.00\\
multi-task baseline   &HRNet18   &51.48  &57.23 &83.43  &14.10 &69.76 &-6.77\\

%\noalign{\smallskip}
%\hdashline
%\noalign{\smallskip}
PAD-Net~\cite{Pad-net_2018}    &HRNet18  &53.60 &59.60 &65.80  &15.3 &72.50 &-4.41\\
ATRC~\cite{atrc_2021}          &HRNet18  &57.89 &57.33 &83.77 &13.99 &69.74 &-4.45\\ 
MQTransformer\cite{mqtrans_xy} &HRNet18  &58.91 &57.43 &83.78 &14.17 &69.80 &-4.20\\
\rowcolor{mygray}DeMT (Ours)   &HRNet18  &59.23	&57.93 &83.93 &14.02 &69.80 &-3.79\\
\hdashline
single task baseline           &Swin-T   &67.81	 &56.32  &82.18  &14.81  &70.90 &0.00\\
multi-task baseline            &Swin-T   &64.74	 &53.25  &76.88  &15.86  &69.00 &-3.23\\
MQTransformer\cite{mqtrans_xy} &Swin-T   &68.24	 &57.05	 &83.40  &14.56  &71.10 &1.07\\
\rowcolor{mygray}DeMT (Ours)   &Swin-T   &69.71	 &57.18  &82.63  &14.56  &71.20 &1.75\\
\hdashline
single task baseline           &Swin-S   &70.83  &59.71  &82.64  &15.13  &71.20 &0.00\\
multi-task baseline            &Swin-S   &68.10  &56.20  &80.64  &16.09  &70.20 &-3.97\\
MQTransformer\cite{mqtrans_xy} &Swin-S   &71.25  &60.11  &84.05  &14.74  &71.80 &1.27\\
\rowcolor{mygray}DeMT (Ours)   &Swin-S   &72.01  &58.96  &83.20  &14.57  &72.10 &1.36\\
%\noalign{\smallskip}
\hdashline
%\noalign{\smallskip}
single task baseline          &Swin-B  &74.91  &62.13 &82.35  &14.83  &73.30  &0.00\\
multi-task baseline           &Swin-B  &73.83  &60.59 &80.75  &16.35  &71.10  &-3.81\\
\rowcolor{mygray}DeMT (Ours)  &Swin-B  &75.33  &63.11 &83.42  &14.54  &73.20  &1.04\\
\hline
  \end{tabular} %}
\end{table*}

\noindent
\textbf{NYUD-v2.}
The Comparisons with state-of-the-art models on the NYUD-v2 dataset are shown in Table~\ref{tab:stoa_nyud_v2}.
We first report results comparison with three different backbones: HRNet18, Swin-T, and Swin-S. 
We demonstrate simultaneous performance improvements over prior work in having smaller parameters, a smaller number of GFLOPs, and better semantic segmentation, depth estimation, surface normal and boundary detection accuracies.
For example, a performance comparison between MQTransformer and DeMT proves the effectiveness of our framework. Besides this, DeMT also consistently outperforms previous state-of-the-art Transformer-based models, such as ATRC~\cite{atrc_2021} and MQTransformer~\cite{mqtrans_xy}.
In addition, we also observe that using a transformer as a backbone model is more promising compared to CNN as the backbone.
Because Transformer-based and CNN-based models use similar GFLOPs, the former shows higher accuracy in all metrics.
Our DeMT obtains 46.36 SemSeg accuracy, which is 6.3\% higher than that of MQTransformer with the same Swin-T backbone and slightly lower FLOPs (100.7G vs. 106.02G).
MuIT~\cite{transfmult2022} reports a 13.3\% and 8.54\% increase in relative performance for semantic segmentation and depth tasks. While we have a 14.74\% and 9.43\% increase. The comparison results show our model also achieves good performance, evaluating the flexibility of our model.
By comparison, our DeMT achieves new records on the NYUD-v2, which are remarkably superior to previous CNNs and Transformers models in terms of all metrics.

\noindent
\textbf{PASCAL-Context.}
We also evaluate the proposed DeMT on PASCAL-Context with three backbones: HRNet18, Swin-T, Swin-S, and Swin-B. 
Table~\ref{tab:stoa_pascal} shows the comparison results.
Our model obtains significantly better results when compared with the baseline and other models. 
For example, DeMT improves MQTransformer~\cite{mqtrans_xy} with the same Swin-T backbone by 1.47 point in SemSeg.
Our DeMT achieves the best performance among models on several metrics and can reach a high performance of 75.33 in the SemSeg task. %which is significant due to the fact that this benchmark is very competitive.

\begin{table*}[!t]
\begin{center}
%\linespread{0.8}
\scriptsize
	\centering
	\caption{Ablation studies and analysis on NYUD-v2 dataset using a Swin-T backbone. Deformable mixer (DM), task interaction (TI) block, and task query (TQ) block are the parts of our model. HR48 denotes HRNet48. The notation ‘$\downarrow$’: lower is better. The notation ‘$\uparrow$’: higher is better. The \textbf{w/} indicates \textbf{"with"}.}
	%\vspace{-1.5mm}
	\setlength{\tabcolsep}{5.pt}
	\subfloat[\small Ablation on components]{
        \label{tab:module}
        \resizebox{0.460\textwidth}{!}{
		\begin{tabularx}{6.4cm}{c|c|c|c|c} 
		       \toprule%[0.15em]
		       \multirow{2}*{Model}    & {SemSeg}   &Depth & Normal & Bound\\
                 & (IoU)$\uparrow$   &(rmse)$\downarrow$   & (mErr)$\downarrow$  &(odsF)$\uparrow$\\
    		  %$N$ & {SemSeg}   &Depth & Normal & Bound\\ %&  & PQ \scriptsize{(Things)} & PQ \scriptsize{(Stuff)}
    	 	\toprule%[0.15em]
    	             baseline   &38.78    &0.6312   &21.05   &75.6\\
         \textbf{w/} DM         &42.40	  &0.6069   &20.83   &76.2\\
         \textbf{w/} DM+TI      &44.44	  &0.5969   &20.75   &76.4\\
         \textbf{w/} DM+TI+TQ   &46.36    &0.5871   &20.65   &76.9\\
			\bottomrule%[0.1em]
		\end{tabularx} }
    } \hfill
    %\vspace{-1.3mm}
    \setlength{\tabcolsep}{8.pt}
    \subfloat[\small Ablation on the depths ($d$)]{
        \label{tab:depth}
        \resizebox{0.460\textwidth}{!}{
		\begin{tabularx}{6.4cm}{c|c|c|c|c} 
		        				\toprule%[0.15em]
    		    \multirow{2}*{$d$} & {SemSeg}   &Depth & Normal & Bound\\ %&  & PQ \scriptsize{(Things)} & PQ
    		   & (IoU)$\uparrow$   &(rmse)$\downarrow$   & (mErr)$\downarrow$  &(odsF)$\uparrow$\\
    		 
    	 	\toprule%[0.15em]
    	 
         1  &46.36    &0.5871   &20.65     &76.9\\
         2  &46.90    &0.5622   &20.05     &77.0\\
         4  &47.71	  &0.5613   &19.90     &77.1\\
         8  &47.16	  &0.5518   &19.87	   &77.1\\         
			\bottomrule%[0.1em]
		\end{tabularx}}
    } \hfill
    %\vspace{-1.3mm}
    \setlength{\tabcolsep}{4.pt}
    \subfloat[\small Ablation on scales]{
        \label{tab:scales}
        \resizebox{0.460\textwidth}{!}{
		\begin{tabularx}{6.4cm}{c|c|c|c|c} 
		        				\toprule%[0.15em]
    		  \multirow{2}*{Scale} & {SemSeg}   &Depth & Normal & Bound\\ %&  & PQ
    		  	   & (IoU)$\uparrow$   &(rmse)$\downarrow$   & (mErr)$\downarrow$  &(odsF)$\uparrow$\\

    	 	\toprule%[0.15em]
             1/4                       &7.51      &1.1961    &33.26    &66.1\\
             1/4, 1/8                  &12.85	  &1.0433    &27.66    &70.4\\
             1/4, 1/8, 1/16            &40.32     &0.6966    &21.44    &76.3\\
             1/4, 1/8, 1/16, 1/32      &46.36     &0.5871    &20.65    &76.9\\
			\bottomrule%[0.1em]
		\end{tabularx}}
    } \hfill
    %\vspace{-1.3mm}
    \subfloat[\small Ablation on backbones]{
        \setlength{\tabcolsep}{5.pt}
        \label{tab:backbone}
        \resizebox{0.460\textwidth}{!}{
	    \begin{tabularx}{6.4cm}{c|c|c|c|c}
		 \toprule%[0.15em]
    		 \multirow{2}*{Backbone}  & {SemSeg}   &Depth & Normal & Bound\\ 
    		 & (IoU)$\uparrow$   &(rmse)$\downarrow$   & (mErr)$\downarrow$  &(odsF)$\uparrow$\\
    		\toprule%[0.15em]
    	       HR48 baseline          &41.96   &0.5543  &20.36   &77.6  \\
    	       HR48 \textbf{w/} ours  &43.84   &0.5517  &19.88   &77.7  \\
    	       %\hdashline
    	       Swin-B baseline         &51.44  &0.5813  &20.44   &77.0 \\
    	       Swin-B \textbf{w/} ours &54.34  &0.5209  &19.21   &78.5  \\
        	\bottomrule%[0.1em]
	    \end{tabularx}}
    } \hfill
    %\vspace{-1.3mm}
\end{center}
\end{table*}

\subsection{Ablation Studies}
We ablate DeMT to understand the contribution of each component and setting using Swin-T on NYUD-v2 dataset.

\noindent
\textbf{Ablation on modules.}
The DeMT model consists of three components: deformable mixer, task interaction, and task query blocks.
As shown in Table~\ref{tab:module}, we demonstrate the advantages of the deformable mixer, task interaction, and task query blocks.
We observe that task interaction block has more effect on the performance, and it is essential to interact the whole task features for task interaction information.
This indicates that task interaction and task query blocks are essential to the task-aware transformer decoder.
From the Figure~\ref{fig:vis_diff_module} and Table~\ref{tab:module} it can be observed that different components are playing a beneficial role.

\noindent
\textbf{Ablation on the depths $d$.}
As shown in Figure~\ref{fig:models}, the depth $d$ is the number of repetitions of the deformable mixer.
We add the $d$ to analyze the effect of the depth of the deformable mixer on the DeMT model.
In Table~\ref{tab:depth}, We vary the number of used deformable mixer depth ($e.g.,$ 1, 2, 4, 8) and compare their performances.
Comparing the first to last row in Table~\ref{tab:depth}, we observe the best performance when the depth is set to 4.
However, as increasing the depth, the parameters and GFLOPs also become more extensive.
Practically, we choose a depth $d$ = 1 for all models in this paper.

\noindent
\textbf{Ablation on scales.}
We explore the influence of using different scale features.
The backbone outputs four-scale (1/4, 1/8, 1/16, 1/32) features.
Table~\ref{tab:scales} shows the influence of using a different number of scales.
Note that the model performance increases obviously with the increasing number of scales.
Our method can capture valuable semantic information for multiple tasks. 
Practically, we choose four-scale features for all models in this paper.

\noindent
\textbf{Ablation on backbones.}
Table~\ref{tab:backbone} shows the results using the different backbones.
To deeper explore the capacity of the our DeMT, we employ extensive backbones to conduct the ablation experiment.
It is worth noting that our DeMT leads to the best performance on all metrics when using Swin-B on NYUD-v2.
In addition, we also observe the inspiring fact that using a larger transformer backbone can easily reach top-tier performance.
The different backbones are compared to demonstrate the generalization of our method.

\subsection{Visualization}
\begin{figure}[!t]
\centering
  \includegraphics[width=0.5\textwidth]{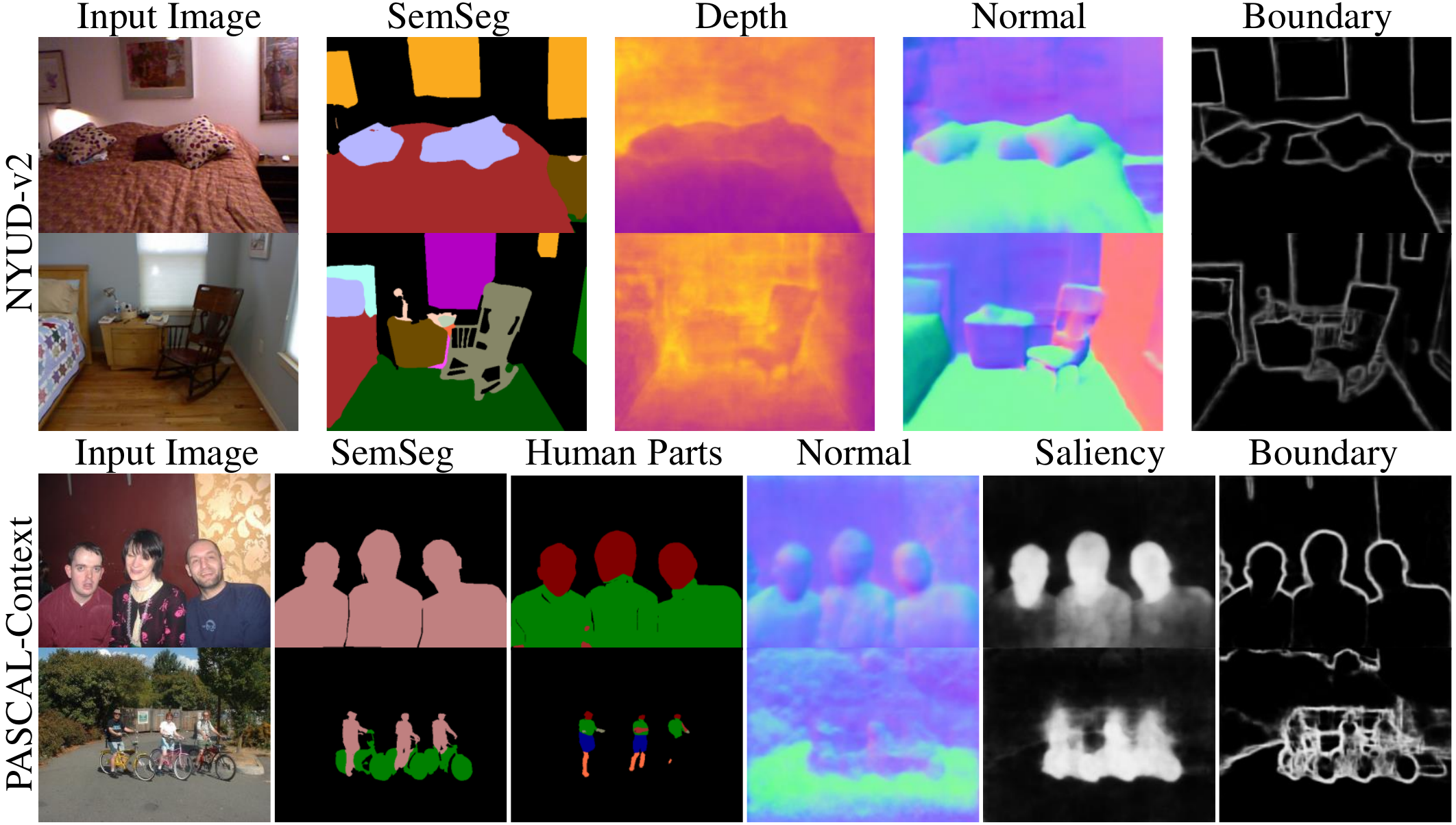}
  \caption{\small The first two rows of the visualization illustrate two examples from the NYUD-v2. The last two rows of the visualization illustrate two examples from the PASCAL-Context.
  }
  \label{fig:vis_data}
\end{figure}

\begin{figure}[!t]
\centering
  \includegraphics[width=0.5\textwidth]{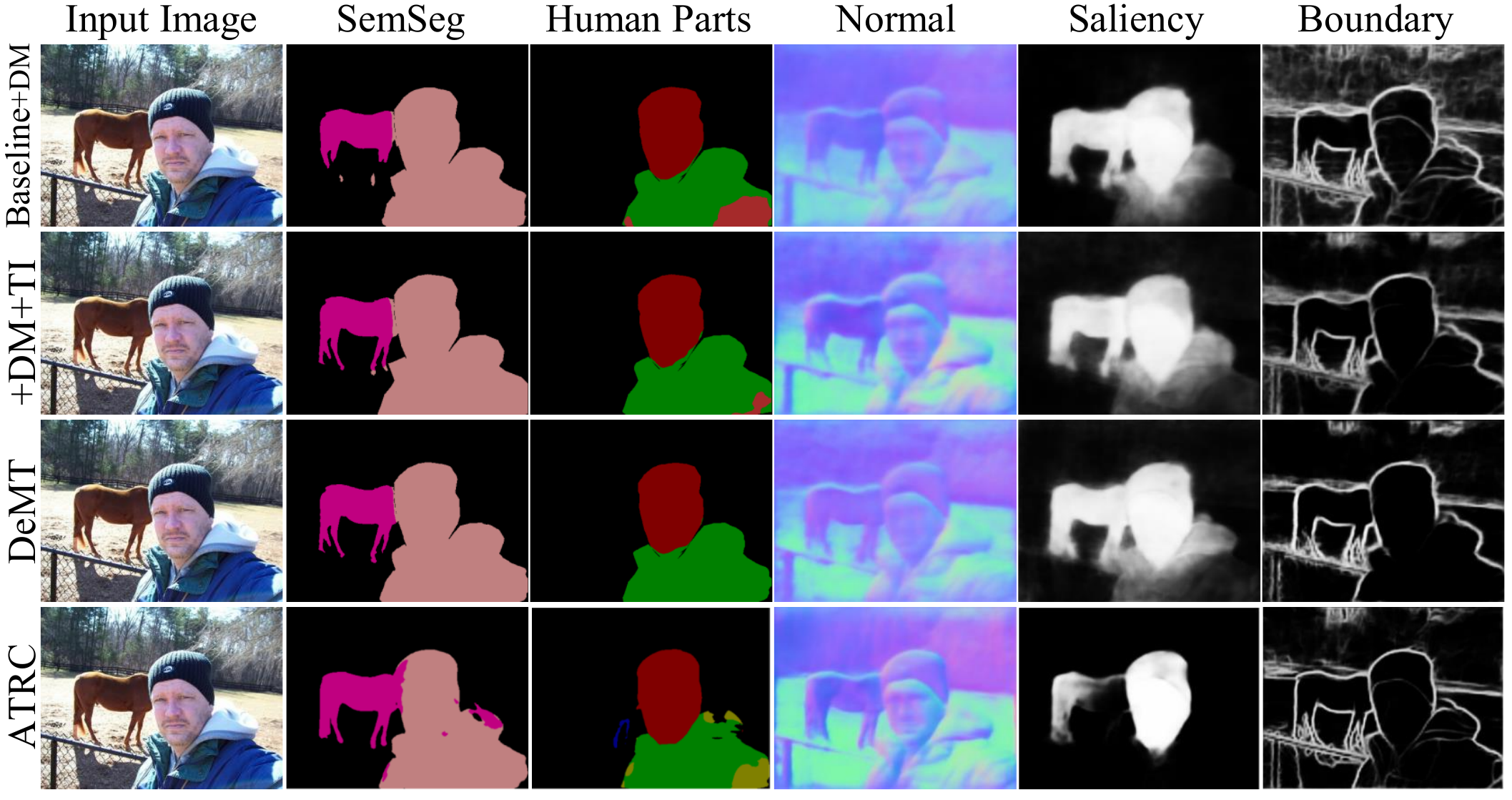}
  \caption{\small Qualitative analysis of the components on PASCAL-Context.
  Visualizations show the components in Table~\ref{tab:module}.
  The last row shows the ATRC model visualization results as a comparison. 
}
  \label{fig:vis_diff_module}
\end{figure}

To deeper understand our DeMT model, we visualize the multiple task predictions.
We show the qualitative results in different dimensions.
For visual analysis (see Figure~\ref{fig:vis_data} and Figure~\ref{fig:vis_diff_module}), we employ a trained model with Swin-T.
Figure~\ref{fig:vis_data} shows the capability of DeMT with Swin-T backbone to perform dense predictions with strong expressive power and successfully capture the task-specific features.
As illustrated in Figure~\ref{fig:vis_data} (last two rows), the second and third columns focus mainly on specific semantics such as human, animal, and other objects. 
Figure~\ref{fig:vis_diff_module} showcases the impact of our approach using different components: while only the deformable mixer encoder fails to visualize some objects, the third row shows DeMT’s improvements to multiple task predictions.
Note that we not only report these results for qualitative understanding of the model but also evaluate it quantitatively in Table~\ref{tab:module}.
We compared the prediction results of the DeMT model with the ATRC (Figure~\ref{fig:vis_diff_module} last row), and our results are significantly better than ATRC, especially on SemSeg and Human Parts tasks.
Our DeMT model produces higher-quality predictions than both the Swin baseline and the existing CNN-based MTL model.

\section{Conclusion}
\label{sec:conclusion}

In this work, we introduce DeMT, a simple and effective method that leverages the combination of both merits of deformable CNN and query-based Transformer for multi-task learning of dense prediction.
Significantly, the deformed feature produced by the deformable mixer encoder is leveraged as a task query in the task-aware transformer decoder to disentangle task-specific features.
Extensive experiments on dense prediction datasets ($i,e.,$ NYUD-v2 and PASCAL-Context) validate the effectiveness of our DeMT model.

\noindent
\textbf{Limitations and future work.}
This work only uses a naive operation to aggregate multi-scale features and could be further improved in two aspects:
considering using the FPN or FPN variant to aggregate multi-scale features and how to design flexible attention to learn more valuable information. 

\section*{Acknowledgements}
This work was done when Yangyang Xu was a research intern at JD Explore Academy.
This work was supported by the National Natural Science Foundation of China under Grants 62122060, 62076188, and the Special Fund of Hubei Luojia Laboratory under Grant 220100014.

% Use \bibliography{yourbibfile} instead or the References section will not appear in your paper
\bibliography{aaai23}

\end{document}